\documentclass[9pt,a4paper,twoside]{rho-class/rho}
\usepackage[english]{babel}
\usepackage{algorithm}     %
\usepackage{algorithmicx}   %
\usepackage{algpseudocode}
\usepackage{amsmath}       %

\setbool{rho-abstract}{true} %
\setbool{corres-info}{true} %

\journalname{Technical Report}
\title{A Vision-Based Analysis of Congestion Pricing in New York City}

\author[1]{Mehmet Kerem Turkcan}
\author[2]{Jhonatan Tavori}
\author[3]{Javad Ghaderi}
\author[4]{Gil Zussman}
\author[5]{Zoran Kostic}
\author[6]{Andrew Smyth}

\affil[1,2,3,4,5,6]{Columbia University, New York, USA}

\leadauthor{Turkcan et al.}
\footinfo{ }
\smalltitle{Technical Report}
\institution{A Vision-Based Analysis of Congestion Pricing in New York City}

\corres{\url{https://mkturkcan.github.io/congestionpricing/}}
\doi{\url{https://www.doi.org/10.57967/hf/4448}}

\begin{abstract}
    We examine the impact of New York City's congestion pricing program through automated analysis of traffic camera data. Our computer vision pipeline processes footage from over 900 cameras distributed throughout Manhattan and New York, comparing traffic patterns from November 2024 through the program's implementation in January 2025 until January 2026. We establish baseline traffic patterns and identify systematic changes in vehicle density across the monitored region. 
\end{abstract}

\keywords{computer vision, urban planning, congestion pricing, traffic analysis, smart cities, urban mobility}

\begin{document}
	
    \maketitle
    \thispagestyle{firststyle}

\section{Introduction}
\rhostart{U}rban congestion pricing has emerged as a key policy tool for managing traffic flow in major metropolitan areas, following successful implementations in Singapore (1975), London (2003), Stockholm (2006), Milan (2008), and Gothenburg (2013) \cite{lehe2019downtown}. On November 14, 2024, New York City announced its own congestion pricing program, effective January 5, 2025, marking the first such implementation in a North American city \cite{MTA_Congestion_Relief_Zone}. Although previous studies have examined congestion pricing effects through manual traffic counts, sensor networks, or limited camera deployments, a visual analysis of city-wide traffic patterns remains challenging due to the scale of data collection and processing needed.

We address this challenge by developing a computer vision pipeline that leverages New York City's existing network of 910 traffic cameras provided by \url{webcams.nyctmc.org} \cite{nyctmc_webcams}. Our methodology enables systematic measurement of traffic density changes both within and around the designated Congestion Relief Zone, while addressing technical challenges inherent in processing low-resolution traffic camera feeds at scale. This approach offers three key contributions:

\begin{enumerate}
    \item A single-stage object detection model (YOLO-LR) optimized for processing low-resolution ($352\times240$) traffic camera footage available to the public, achieving improved detection accuracy compared to standard high-resolution models;
    \item A scalable infrastructure for real-time processing of hundreds of simultaneous video feeds using distributed computing;
    \item A methodology for analyzing traffic pattern changes that accounts for temporal variations and systematic biases in camera-based measurements.
\end{enumerate}

By establishing this framework, we enable quantitative evaluation of large-scale traffic policy interventions using existing urban camera infrastructure. Our analysis provides insights into the effects of the implementation of congestion pricing and the spatial redistribution of traffic patterns throughout the metropolitan area.

\begin{algorithm}[!t]

\caption{Traffic Pattern Analysis}
\label{alg:pattern_analysis}
\label{alg:1}
\begin{algorithmic}[1]
\Require
    \State $D = \{d_1, ..., d_n\}$ \Comment{Raw traffic count observations}
    \State $t = \{t_1, ..., t_n\}$ \Comment{Timestamps for each observation}
    \State $S = \{s_1, ..., s_m\}$ \Comment{Set of traffic count sources}
    \State $w$ \Comment{Rolling window size}
    \State $T_{split}$ \Comment{Split timestamp}

\Procedure{ProcessTrafficData}{$D, t, S, w, T_{split}$}
    \For{each source $s \in S$}
        \State $\bar{D}_s \gets \text{RollingMean}(D_s, w)$ \Comment{Smooth counts for source s}
    \EndFor
    
    \State $H \gets \{0,...,23\}$ \Comment{Hours of day}
    \State $W \gets \{\text{weekday}, \text{weekend}\}$ \Comment{Day types}
    \State $P \gets \{\text{before}, \text{after}\}$ \Comment{Split periods}
    
    \For{each source $s \in S$}
        \For{each $(h,w,p) \in H \times W \times P$}
            \State $D_{s,h,w,p} \gets \{d_i \in \bar{D}_s | \text{Hour}(t_i)=h, \text{DayType}(t_i)=w, t_i \in p\}$
            \State $\mu_{s,h,w,p} \gets \text{Mean}(D_{s,h,w,p})$
        \EndFor
    \EndFor
    
    \For{each time window $[h_1, h_2] \in \text{TimeWindows}$}
        \For{each $(s,w) \in S \times W$}
            \State $\text{Peak}_{s,w,before} \gets \max_{h \in [h_1,h_2]} \mu_{s,h,w,before}$
            \State $\text{Peak}_{s,w,after} \gets \max_{h \in [h_1,h_2]} \mu_{s,h,w,after}$
            \State $\Delta_{peak,s,w} \gets \text{Peak}_{s,w,after} - \text{Peak}_{s,w,before}$
        \EndFor
    \EndFor
    
    \State \Return $\{\Delta_{peak,s,w} | s \in S, w \in W\}$ \Comment{Peak differences for each source}
\EndProcedure

\State \textbf{where:}
\State $\quad \text{TimeWindows} = \{[0,23], [6,9], [9,15], [15,18]\}$ \Comment{Day, Morning, Midday, Afternoon}
\State $\quad \text{RollingMean}(X,w) = \frac{1}{w}\sum_{i=0}^{w-1} x_{t-i}$ \Comment{w-point moving average}
\end{algorithmic}

\end{algorithm}

\begin{table*}[t!]
    \centering
    \caption{Performance comparison of YOLO models across different classes in COCO val.}
    \label{tab:detection_results}
    \begin{tabular}{l c | ccc | ccc | ccc}
        \toprule
        Class & Instances & \multicolumn{3}{c|}{YOLO-LR (imgsz=352)} & \multicolumn{3}{c|}{YOLO11n (imgsz=352)} & \multicolumn{3}{c}{YOLO11n (imgsz=640)} \\
        \cmidrule(lr){3-5} \cmidrule(lr){6-8} \cmidrule(lr){9-11} \cmidrule(lr){9-11}
        & & R & mAP50 & mAP50-95 & R & mAP50 & mAP50-95 & R & mAP50 & mAP50-95 \\
        \midrule
        All        & 3296 & 0.488 & 0.537 & 0.357 & 0.420 & 0.467 & 0.305 & 0.523 & 0.597 & 0.413 \\
        Bicycle    &  314 & 0.341 & 0.387 & 0.227 & 0.277 & 0.332 & 0.188 & 0.392 & 0.478 & 0.280 \\
        Car        & 1918 & 0.481 & 0.513 & 0.305 & 0.388 & 0.410 & 0.237 & 0.523 & 0.582 & 0.375 \\
        Motorcycle &  367 & 0.537 & 0.605 & 0.373 & 0.512 & 0.577 & 0.338 & 0.585 & 0.672 & 0.440 \\
        Bus        &  283 & 0.678 & 0.740 & 0.595 & 0.601 & 0.652 & 0.537 & 0.707 & 0.772 & 0.647 \\
        Truck      &  414 & 0.403 & 0.437 & 0.287 & 0.321 & 0.363 & 0.224 & 0.408 & 0.480 & 0.326 \\
        \bottomrule
    \end{tabular}
\end{table*}

\section{Algorithm}

\subsection{Object Detection}
We address the technical challenges of processing low-resolution traffic camera footage by developing YOLO-LR, an adaptation of YOLO optimized for low-resolution input. Of the 910 cameras in our dataset, 770 (84.6\%) output footage at $352\times240$ pixels, significantly below standard object detection dataset resolutions. We train our model on the COCO dataset at this resolution, specifically focusing on five relevant classes: bicycle, car, motorcycle, bus, and truck. Images are resized and padded to $352\times352$ resolution. We show comparative evaluation metrics in Table~\ref{tab:detection_results} against the standard YOLO model trained at $640\times640$ resolution. Our results show that the YOLO-LR model demonstrates improved performance on low-resolution traffic footage.

\subsection{Data Collection Infrastructure}
Our data collection system employs a distributed architecture with 16 parallel workers for frame capture and processing. The pipeline processes captured frames in batches of 64 using the YOLO-LR model compiled using TensorRT, with automatic discarding of frames exceeding a 100ms download threshold. All processing is performed on a single compute node equipped with dual NVIDIA A100 40GB GPUs. Detection results are stored in a SQLite database for subsequent analysis. All frames are discarded after processing.

\subsection{Pattern Analysis}
Our pattern analysis methodology addresses three key challenges in urban traffic analysis: temporal variability, spatial heterogeneity, and measurement noise. Algorithm~\ref{alg:pattern_analysis} details our approach to quantifying traffic pattern changes through a multi-scale temporal analysis framework. 

To quantify congestion dynamics, we introduce Peak Hour Differentials (PHD), a metric that captures intervention-induced changes in vehicle densities, while accounting for temporal dependencies and systematic biases in camera-based observations.

Given a traffic source \( s \) and time \( t \), let \( D_s(t) \) denote the vehicle count at time \( t \), and define the rolling mean vehicle density over a window of size \( \omega \) as:
\begin{equation}
    \bar{D}_s(t) = \frac{1}{\omega} \sum_{i=0}^{\omega-1} D_s(t - i),
\end{equation}
To assess congestion changes before and after intervention, we define the time-partitioned expectation of vehicle density for a given source \( s \), hour \( h \), and day type \( w \) as:
\begin{equation}
    \mu_{s,h,w,p} = \mathbb{E} \big[ D_s(t) \mid \text{Hour}(t) = h, \text{DayType}(t) = w, t \in p \big],
\end{equation}
where \( p \) denotes either the pre- or post-intervention period.

For each source \( s \) and day type \( w \), we compute peak densities over specified time windows \( [h_1, h_2] \), where
\begin{equation}
    \text{Peak}_{s,w,p} = \max_{h \in [h_1, h_2]} \mu_{s,h,w,p}.
\end{equation}
The Peak Hour Differential (PHD), measuring congestion shifts post-intervention, is then given by
\begin{equation}
    \text{PHD}_{s,w} = \text{Peak}_{s,w,\text{after}} - \text{Peak}_{s,w,\text{before}}.
\end{equation}
This metric provides three key advantages: (i) it directly quantifies changes in peak congestion, the primary target of pricing interventions, while maintaining clear physical interpretation in terms of vehicle density, (ii) by computing peaks within defined windows rather than at fixed times, it captures potential shifts in peak timing induced by the intervention, and (iii) uses a rolling mean to counter short-term fluctuations and anomalies.

\section{Limitations}

Our approach has some important limitations: (i) stationary vehicles are included in traffic density calculations, which may affect measurements in areas with high street parking density, (ii) the baseline comparison period (starting mid-November) provides limited seasonal context, (iii) the current implementation analyzes aggregate traffic flow without distinguishing between individual lanes or travel directions, (iv) camera-based instantaneous vehicle counts serve as a proxy measure and may not directly correlate with actual transit times or congestion levels, (v) New York experienced unusually cold weather during January-February in 2025, which could have affected commuter behavior significantly.

\printbibliography

\end{document}